\documentclass[10pt,twocolumn,letterpaper]{article}

\usepackage{iccv}
\usepackage{times}
\usepackage{epsfig}
\usepackage{graphicx}
\usepackage{amsmath}
\usepackage{amssymb}
\usepackage[accsupp]{axessibility}
\usepackage[symbol]{footmisc}

\usepackage{svg}
\usepackage{amsmath}

\DeclareMathOperator*{\argmin}{arg\,min}

\newcommand{\mat}[1]{\mathbf{#1}}
\newcommand{\set}[1]{\mathcal{#1}}
\newcommand{\vect}[1]{\mathbf{#1}}

\newcommand{\pose}[0]{\theta}

\newcommand{\keyfeature}[0]{NSF~}
\usepackage{epsfig}
\usepackage{graphicx}
\usepackage{comment}
\usepackage{amsmath,amssymb} % define this before the line numbering.
\usepackage{xcolor}
\usepackage[export]{adjustbox}
\usepackage{bbm}
\usepackage{overpic}
\usepackage{times}

\def\etal{{\em et al. }}
\usepackage[font=small,skip=5pt]{caption}
\usepackage{float}

\usepackage{multirow}
\usepackage{tabularx}
\usepackage{soul}
\usepackage{array}
\usepackage{comment}
\usepackage{outlines}

\usepackage[mathscr]{euscript}
\newcolumntype{L}[1]{>{\raggedright\let\newline\\arraybackslash\hspace{0pt}}m{#1}}
\newcolumntype{C}[1]{>{\centering\let\newline\\arraybackslash\hspace{0pt}}m{#1}}
\newcolumntype{R}[1]{>{\raggedleft\let\newline\\arraybackslash\hspace{0pt}}m{#1}}
\newcolumntype{Y}{>{\centering\arraybackslash}X}
\usepackage{booktabs}
\usepackage{enumitem}

\usepackage{amsmath}

\usepackage[pagebackref=true,breaklinks=true,colorlinks,bookmarks=false]{hyperref}

\iccvfinalcopy % *** Uncomment this line for the final submission

\def\iccvPaperID{1052} % *** Enter the ICCV Paper ID here

\ificcvfinal\fi

\begin{document}

%%%%%%%%% TITLE
\title{NSF: Neural Surface Fields for Human Modeling from Monocular Depth}

\author{
    Yuxuan Xue\textsuperscript{1,2,*}, 
    Bharat Lal Bhatnagar\textsuperscript{1,2,3,4,*},
    Riccardo Marin\textsuperscript{1,2},
    Nikolaos Sarafianos\textsuperscript{4},  
    Yuanlu Xu\textsuperscript{4},\\
    Gerard Pons-Moll\textsuperscript{1,2,3,$\dagger$},
    Tony Tung\textsuperscript{4,$\dagger$},
\\[0.4ex]
\textsuperscript{1~}Tübingen AI Center\quad
	\textsuperscript{2~}University of Tübingen\quad \\
 \textsuperscript{3~}Max Planck Institute for Informatics\quad
        \textsuperscript{4~}Meta Reality Labs Research\quad \\ 
        \quad \textsuperscript{$\dagger$~}Project Lead \\
{\tt\scriptsize \{yuxuan.xue, riccardo.marin, gerard.pons-moll\}@uni-tuebingen.de} \\ {\tt\scriptsize \{bharatbhatnagar, nsarafianos, yuanluxu, tony.tung\}@meta.com}
}

\makeatletter
\let\@oldmaketitle\@maketitle%
\renewcommand{\@maketitle}{
	\@oldmaketitle%
	\begin{center}
 	\includegraphics[width=1\linewidth]{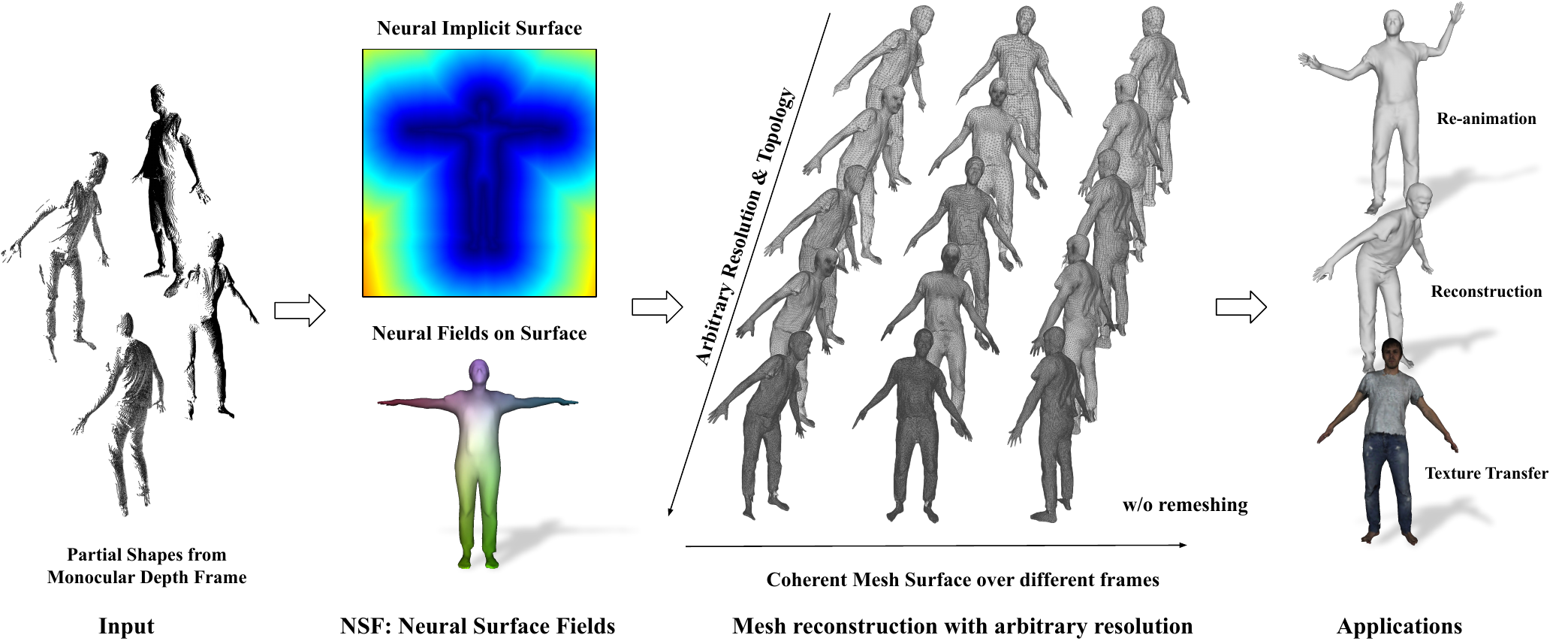}

	\end{center}
    \refstepcounter{figure}\normalfont \footnotesize
Figure~\thefigure. Given a sequence of monocular depth frames of a person, NSF learns a detailed clothed body model of the person. The clothed body model is controllable and can be used for reconstruction and re-animation at arbitrary mesh resolution while maintaining the coherency whithout retraining.
% Our body model is controllable and can be used for reconstruction, re-animation and can even handle textures.
	\label{fig:teaser}
\vspace{5mm}
}

\makeatother
\maketitle
\ificcvfinal\thispagestyle{empty}\fi

\begin{abstract}

Obtaining personalized 3D animatable avatars from a monocular camera has several real world applications in gaming, virtual try-on, animation, and VR/XR, etc. 
However, it is very challenging to model dynamic and fine-grained clothing deformations from such sparse data.
Existing methods for modeling 3D humans from depth data have limitations in terms of computational efficiency, mesh coherency, and flexibility in resolution and topology. 
For instance, reconstructing shapes using implicit functions and extracting explicit meshes per frame is computationally expensive and cannot ensure coherent meshes across frames. Moreover, predicting per-vertex deformations on a pre-designed human template with a discrete surface lacks flexibility in resolution and topology.
To overcome these limitations, we propose a novel method `\keyfeature: Neural Surface Fields' for modeling 3D clothed humans from monocular depth. 
NSF defines a neural field solely on the base surface which models a continuous and flexible displacement field. NSF can be adapted to the base surface with different resolution and topology without retraining at inference time.
Compared to existing approaches, our method eliminates the expensive per-frame surface extraction while maintaining mesh coherency, and is capable of reconstructing meshes with arbitrary resolution without retraining.
To foster research in this direction, we release our code in project page at: \href{https://yuxuan-xue.com/nsf}{https://yuxuan-xue.com/nsf}.
\footnotetext{* denotes equal contribution}

\end{abstract}
\section{Introduction}\label{sec:introduction}

Human modeling is an active and challenging field of research that has applications in Computer Vision and Graphics.
Recent advancements in data acquisition techniques~\cite{huang2018dip,saito2019pifu, saito2020pifuhd, xiu2022econ, xiu2022icon, xue2022events, nguyen2021human} have opened new opportunities for capturing and digitising human appearance. 
Building digital avatars has found applications in behavioural studies~\cite{dollinger2023embodied, fiedler2023embodiment, guzov2021HPS, huang2022intercap, liu2023gshell, petrov2023popup, xie22chore, xie2023vistracker} %, medical~\cite{zhang2022secure, sun2022metaverse} studies as well.
and generative modelling~\cite{huang2023tech, liao2023tada, Liu2023MeshDiffusion, qiu2023oft}.
Our goal is to build body model which is controllable i.e., animatable with different poses, and detailed i.e. it should faithfully produce details such as garments wrinkles under different poses.

In recent years, researchers have looked into learning clothed human models from full sequences of 4D scans~\cite{chen2021snarf, lin2022fite, ma2022skirt, ma2021pop, shunsuke2021scanimate, tiwari21neuralgif}. 4D scans provide rich information about the subject appearance, but they also require exclusive technology, pre-processing, and expert intervention at times, which makes this difficult to scale.
A more user friendly line relies on the input with monocular depth from devices such as Kinects~\cite{burov2021dsfn, dong2022pina, kim2022lapfusion, yu2017bodyfusion, yu2018doublefusion}. Such data is easier to obtain and already supported by consumer-grade devices. But this flexibility comes at the cost of additional sensor noise, thus complicating the learning process. 

To mitigate the noise in input data, parametric models such as SMPL~\cite{loper2015smpl} and its successors~\cite{alldieck19learning2reconstruct, bhatnagar2019mgn, pavlakos2019smplx, xu2020ghum, bhatnagar2020loopreg}, can provide a good statistical prior for capturing pose and the overall shape of the person. 
Also, relying on a template naturally supports information transfer across subjects and poses. 
However, designing a pipeline around a specific template restricts the expressivity of the model, which makes the methods less flexible (e.g., limited to tight garments). 
A common representation to relax the topology constraints is point clouds~\cite{lin2022fite, ma2022skirt, ma2021pop, zhang2023closet}. Recently, point based neural implicit representations~\cite{chen2021snarf,dong2022pina, shunsuke2021scanimate, tiwari21neuralgif, wang2021metaavatar, bhatnagar2020ipnet} demonstrated incredible expressive power. But many real applications (e.g., animation, texture transfer) require a 3D mesh.
Hence, these approaches require running costly algorithms~\cite{LorensenC87mcubes, kazhdan2013screened} to reconstruct a supporting surface. Extracting a surface for every frame causes a computational burden and also results in inconsistent triangulations, which further complicate downstream tasks. 
Some works~\cite{burov2021dsfn, kim2022lapfusion} address this issue by predicting displacements on SMPL vertices for modeling clothed humans. While these methods yield coherent mesh reconstruction, they are constrained by the resolution and topology of SMPL template.

We pose ourselves the following goal: starting only from a set of partial shapes from monocular depth frames, can we learn a clothed body model that is \emph{flexible} and \emph{coherent} across different frames, with a \emph{limited computational cost for surface extraction}? % \YX{sounds redundant}

To this end, we propose \emph{\keyfeature: Neural Surface Fields}; a neural field defined continuously all over the surface. Given a canonical shape, represented with an implicit function, we use NSF to define a continuous field over the surface, capable of modeling detailed deformations. 
Using NSF, we can reconstruct a \emph{coherent mesh} in the canonical space at any resolution with just one run of surface extraction algorithms, and share it across all the different poses.
This formulation avoids per-frame surface extraction which is $\sim40$x and $\sim 180$x faster compared to point-based works~\cite{lin2022fite, ma2022skirt, ma2021pop, zhang2023closet} using Poisson reconstruction and implicit-based works~\cite{chen2021snarf,dong2022pina, shunsuke2021scanimate, tiwari21neuralgif, wang2021metaavatar} using marching cube at similar resolution, respectively. 
After training, NSF can be adapted to \emph{arbitrary resolutions} at inference time, depending on the application.
This step is possible since NSF is continuously defined all over the surface, and hence it is able to support any discretization. Compared to other feature representations, NSF is more compact, saving  $97.4\%$ of memory compared to volumetric representation and $86.0\%$ compared to triplane features at $128^3$ resolution. 

We validate our self-supervised approach on several datasets~\cite{burov2021dsfn, kim2022lapfusion, li2021posefusion, ma2020cape, ponsmoll2017clothcap},  showing better performance than competitors, even when some of them requires subject-specific training~\cite{burov2021dsfn, dong2022pina, ma2021pop, palafox2021npms, palafox2021spams, wang2021metaavatar}. We show the practical benefits of NSF in shape reconstruction, animation, and texture transfer application, with the flexibility and the coherency that is not attainable for prior works~\cite{burov2021dsfn, dong2022pina}. 

In summary, our contributions can be summarized as:
\begin{itemize}
    \item We propose \emph{\keyfeature: Neural Surface Fields}; a continuous neural field defined over the surface in a canonical space which is compact, efficient, and supports arbitrary mesh discretizations without retraining.
    \item We propose a method to learn an animatable human avatar from a monocular depth sequence; NSF let us recover detailed shape information from monocular depth frames. Our self-supervised approach handles subjects with different clothing geometries and textures. To the best of our knowledge, NSF is the first work in avatarization which directly output mesh at arbitrary resolution while maintaining the coherency across different poses.
\end{itemize}

\section{Related Work}\label{sec:related}

\noindent\textbf{Human Capture.} 
Clothed human reconstruction is a rapidly evolving field of research that aims to create realistic and detailed digital models of humans. 
Recent work~\cite{habermann2022hdhumans, he2021arch++, huang2020arch, saito2019pifu, saito2020pifuhd, xiu2022icon, xiu2022econ, zheng2021pamir} can reconstruct humans from a single RGB image but are not as accurate. 
% Additionally, human reconstruction from monocular depth streams is also a popular topic for long time. 
Methods such as KinectFusion~\cite{newcombe2011kinectfusion} and DynamicFusion~\cite{newcombe2015dynamicfusion} fuse depth measurements over time to create a complete and accurate model. While these are general and not restricted to humans, BodyFusion~\cite{yu2017bodyfusion} and DoubleFusion~\cite{yu2018doublefusion} incorporate priors on human motion and shape, fusing partial depths in real-time to obtain improved reconstruction.
However, these methods are complicated to setup and require expert intervention. Moreover, their code is unavailable.
With the advent of deep learning methods, data-driven methods such as IF-Nets~\cite{chibane20ifnet}, reconstruct humans by learning a prior from a large dataset. IP-Net~\cite{bhatnagar2020ipnet} further fits a parametric model to the implicit reconstruction to make the mesh controllable.
These approaches only capture static humans and do not capture the pose dependent deformations, thus lacking realism.

\noindent\textbf{Implicit Neural Avatar.} 
In the last few years, outstanding results produced by Neural Radiance Field (NeRF)~\cite{mildenhall2021nerf} have motivated scholars to model the clothed human as implicit neural representations. There's a plethora of NeRF-based approaches for humans modeling that provide animatable avatars starting from monocular RGB videos~\cite{feng2022scarf, feng2023disavatar, jiang2022neuman, peng2021animatable, shao2022doublefield, weng_humannerf_2022_cvpr, Zhao2022humannerf}.
Apart from constructing the human model using RGB images, a common and straightforward approach involves learning the implicit neural avatar from geometric data, such as scans ~\cite{chen2022gdna,chen2021snarf, deng2019nasa, Mihajlovic2021leap, tiwari21neuralgif, wang2021metaavatar,zins2021data, tiwari20sizer}. Furthermore, PINA~\cite{dong2022pina} models an implicit personalized avatar using monocular depth sequences, which share the same input as our work. However, it is important to note that these implicit-based methods are \emph{subject-specific} and are unable to model multiple subjects simultaneously.
% MetaAvatar~\cite{wang2021metaavatar} uses MetaLearning to perform the quick adaptation to unseen subjects, but need to train the hypernetwork from full scan. 
% This is a direct effect of nesting the template inside the network, making it hardly accessible. 
Furthermore, these methods that rely on implicit representations utilize neural networks to parameterize the shape, and cannot directly provide explicit meshes as output. In order to obtain a mesh representation, an extensive computation of marching cubes is performed for \emph{each frame}, resulting in computationally expensive operations. Moreover, the extracted surface using marching cubes lacks \emph{coherence} across different frames. This lack of coherency leads to the loss of natural correspondence and poses additional hindrances in applying these methods to downstream tasks, e.g. texture transfer between the input and the learned shape.

\noindent\textbf{Explicit Parameterized Avatars.}
SMPL~\cite{loper2015smpl} is a popular parametric human model. However, it only models the naked body shape and pose, and lacks details.
% is limited to the body statistic captured during its training.
Hence, several extensions have been proposed to add further details like hands~\cite{romero2022embodied}, face~\cite{pavlakos2019smplx}, soft-tissues~\cite{ponsmoll2015dyna} and clothing~\cite{patel2020tailornet, ponsmoll2017clothcap, ponsmoll2015dyna, zhang2017buff, bhatnagar2019mgn}. 
Many works model deformations~\cite{burov2021dsfn, kim2022lapfusion, ma2022skirt, ma2021pop, ma2021scale, zhang2017buff, zhang2023closet, alldieck19learning2reconstruct}, by fitting SMPL model and adding cloth wrinkles as displacement on top of the coarse shape. 
Although they reconstruct coherent shapes, they are often limited by the resolution and topology of the SMPL template, making them less flexible compared to implicit-based methods. 
To overcome this limitation, Lin \etal~\cite{lin2022fite} proposed to learn the fusion shape using implicit occupancy network, which is not constrained by the SMPL topology and can represent loose garments like skirts. However, this approach relies on complete scans and registered mesh data to provide ground-truth occupancy labels.
Moreover, these point-based works~\cite{lin2022fite, ma2022skirt, ma2021pop} need to perform Poisson Reconstruction at each frame to obtain the mesh.
In contrast, our approach fuses monocular raw depth inputs into a canonical space to obtain a coarse, pose-independent base shape without any supervision, which is difficult to obtain from partial shape data. 
We then learn pose-dependent neural surface fields~(Sec.~\ref{sec:nsf}) on top of the coarse shapes, which allow us to obtain detailed shapes at arbitrary resolutions.
In summary, our approach offers flexibility and efficiency in generating coherent meshes, and eliminates the need for Marching Cubes or Poisson Reconstruction at each frame (Sec.~\ref{sec:surface_extraction}).

\section{NSF: Neural Surface Fields}
\label{sec:nsf}

\begin{figure*}[t]
    \centering
    \captionsetup{type=figure}
    \includegraphics[width=\linewidth]{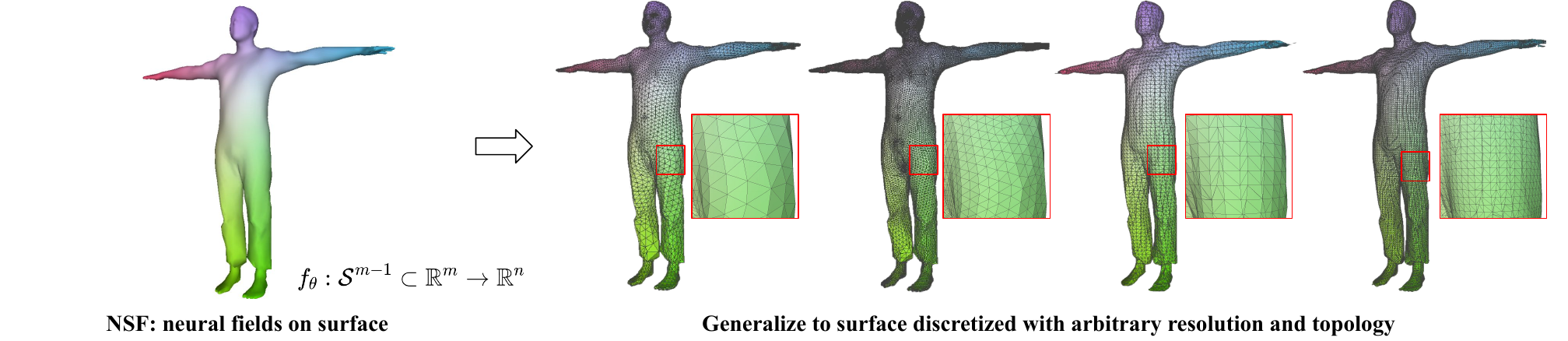}
    \caption{\label{fig:NSF_theory} We show an example of NSF decoding to surface color $\in \mathbb{R}^3$. On the right of arrow shows that NSF can be queried with the surface with arbitrary resolution or topology without retraining.}
\end{figure*}

\paragraph{Neural Fields.} A neural field is a field parametrized by a neural network~\cite{xie2021neuralfield}:
\begin{equation}
f_{\phi}: \mathbb{R}^m \rightarrow \mathbb{R}^n,
\end{equation}
where $\phi$ are the learnable parameters. Neural field defined in Euclidean space $\mathbb{R}^3$ has been widely-used to represent various geometries like distance~\cite{park2019deepsdf}, occupancy~\cite{mescheder_occupacynet}, and radiance~\cite{mildenhall2021nerf} functions, correspondences~\cite{bhatnagar2020ipnet}, contacts~\cite{karunratanakul2020grasping, bhatnagar22behave, zhou2022toch}, parametric body models~\cite{bhatnagar2020loopreg}, and so on.

\paragraph{Neural Surface Fields.} When a field carries information about an object that occupies a limited volume bounded by a 2D surface $\mathcal{S}$, we know in advance that much region of the space will not be ever queried, causing a waste of computational and memory resources~\cite{chibane20ifnet,bhatnagar2020ipnet,bhatnagar2020loopreg}. Following this intuition, we are interested to define the field only on the 2D surface $\mathcal{S}^2$:
\begin{equation}
f_{\phi}: \mathcal{S}^2 \subset \mathbb{R}^3 \rightarrow \mathbb{R}^n.
\end{equation}
We call this representation \emph{Neural Surface Fields (NSF)}. Recent work~\cite{koestler2022intrinsic} defines the neural field with the eigenfunction of the Laplace-Beltrami Operator on the surface, and hence are defined just for a specific discretization of the geometry. Instead, our approach is more general and produces a continuous field independent of the underlying discretization of the object.

Embedding the neural fields on a surface is advantageous due to the ability to combine properties with mesh surface coherency and connectivity as shown in Fig.~\ref{fig:NSF_theory}.
In our work, we leverage NSF to learn a continuous deformation field which models the detailed clothing deformations on the surface of the coarse clothed human shape (Sec.~\ref{sec:manifold_implicit_function}).

\section{NSF for Human Modelling}\label{sec:method_v2}

In this section we show the advantages of NSF by incorporating it into an avatarization method. Before diving into the method details, we will state our goal, define method's input, and provide a general overview.% . and point to the key ideas.

\noindent\textbf{Input.} Let $s=\{1,... N\}$ be the set of subjects. For each subject, our method takes as input a sequence of monocular depth point clouds, $\set{X}^s=\{ \mat{X}^s_1,... \mat{X}^s_{T_s} \}$. Each $\mat{X}_t^s$ is a set of unordered points $\{ \vect{x}^{s,t}_j \}^{L_{s,t}}_{j=1}$ where $L_{s,t}$ represents the number of points in the monocular point cloud at time $t$. Also, for subject sequence we take as input the corresponding 3D poses $\set{\pose}^s=\{ \set{\pose}^s_1,... \set{\pose}^s_{T_s} \}$.

\begin{figure*}[t]
    \centering
    \captionsetup{type=figure}
    \includegraphics[width=\linewidth]{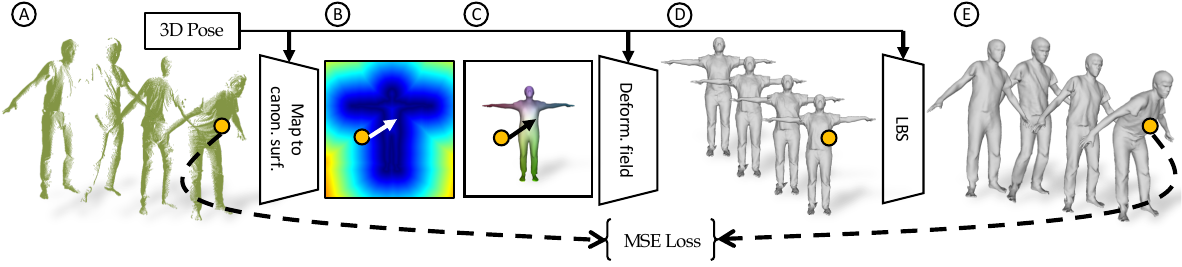}
    \caption{\label{fig:pipeline} We propose a method to learn animatable body models of people from monocular depth point clouds and 3D poses (A). We learn an implicit canonical shape of a person (B) by fusing the partial point clouds. To get fine details, we learn pose-dependent deformations as a continuous field on the surface of the fusion shape (C), using our \emph{neural surface fields}. By predicting deformations in canonical pose (D), we pose our 3D reconstructions using simple LBS (E). Our approach can be trained with self-supervision.}
\end{figure*}

\noindent\textbf{Output.} Our goal is to learn subject-specific body models, $\set{M} = \{{M}^1,...{M}^N\}$. Each model $M^s(\vect{p}, \theta)$ can transform points $\vect{p} \in \mathbb{R}^3$ from a neutral pose in canonical space to the target pose $\theta$, taking the shape and clothing of the subject into account.
Our models are complete, detailed, and contain pose dependent garment deformations of the subject.

% \RM{Are we describing inference or training time?}
\noindent\textbf{Overview.} We kindly ask readers to refer Fig.~\ref{fig:pipeline} for an overview of our method. 
To learn the body model of each subject, (A) we unpose the input point clouds (Sec.~\ref{sec:fusion_shape}) to a neutral pose using inverse skinning, and (B) we fuse them to learn an implicit (SDF) \emph{canonical shape} $\set{B}^s$ (Sec.~\ref{sec:fusion_shape}). Our canonical shape is continuous, and the fusion of different depths averages out fine-grained details generated by the subject poses. On top of our canonical shape, (C) we train NSF (Sec.~\ref{sec:manifold_implicit_function}), which predicts the pose-dependent deformation for each point on the continuous canonical surface, (D) recovering the cloth deformation for a specific pose of the subject (Sec.~\ref{sec:manifold_implicit_function}). Finally, (E) we use LBS to pose the human model (Sec.~\ref{sec:cycle_consistency}). The method is optimized using a cycle-consistency loss between the input point cloud and our predicted shape. 
For simplicity we drop $s$ from subsequent notation and explain our method for a single subject. We will reintroduce $s$ for parts of the manuscript dealing with multiple subjects.

\subsection{Fusion Shape from Monocular Depth}
\label{sec:fusion_shape}
\noindent\textbf{Canonicalization.} 
To build our person-specific canonical shape, we unpose every $\mat{X}_t$ input point cloud to a neutral pose. The corresponding canonical points $\mat{X}^c_t$ for input points can be found using iterative root finding~\cite{chen2021snarf, li2022tava}:
\begin{equation}
    % \set{X}^{c*}, w_i^* = 
    \underset{\mat{X}^c_t, w}{\argmin} \sum_{t=1}^{T} \left( \left( \sum_{i=1}^{K}w(\mat{X}^c_t)_i \cdot \mat{T}_i(\theta_t) \right) \mat{X}^c_t - \mat{X}_t \right).
    % \(\(\sum_{i=1}^{24}w_i(\mat{X}^c)\mat{T}_i\)\mat{X}^c - \mat{X}_t), 
    \label{eq:canonicalization}
\end{equation}
where $K$ is the number of joints, and $w(\cdot)_i$ and $\mat{T}_i$ are the skinning weights and joint transformation for joint $i$ respectively. 
We utilize the iterative root finding in canonicalization together with the pre-diffused SMPL skinning field in FiTE~\cite{lin2022fite} to avoid ambiguous solutions.
We unpose all input observation $\set{X}=\{\mat{X}_t\}_{t=1}^T$ into canonical partial shapes $\set{X}^c = \{ \mat{X}^c_t \}_{t=1}^T$. 

\noindent\textbf{Implicit Fusion Shape.} 
Since the inverse skinning does not account for pose-dependent deformations operates at a human level, the point cloud $\mat{X}^c_t$ resulting from our canonicalization process still contains non-rigid deformation specific to the subject poses. 
To remove the influence of single poses and obtain a coarse canonical shape $\set{B}$, our idea is to fuse every $\{\mat{X}^c_t\}_{t=0}^{T}$ by learning an implicit surface in the canonical space.
Concretely, we represent $\set{B}^s$ as an implicit SDF in~\cite{park2019deepsdf}, composed by a neural network $f^\text{shape}(\cdot|\phi^\text{shape})$ parameterised by parameters $\phi^\text{shape}$, that takes as an input a subject specific latent code $\vect{h}^s \in \mathbb{R}^{256}$ and a query point $\vect{x} \in \mathbb{R}^3$, to predict an SDF value. 
The subject-specific latent codes $\set{H}=\{\vect{h}^s\}_{s=1}^N$, and the decoder parameters $\phi^{shape}$, are optimised with the self-supervised objective~\cite{gropp2020igr} below:
\begin{equation}
    \label{eq:sdf}
    E^\text{shape}(\phi^\text{shape},\set{H}) = E_\text{geo} + \lambda_1 E_\text{eik} % + \lambda_2 E_\text{reg},
\end{equation}
\begin{multline}
    E_\text{geo}(\phi^\text{shape},\set{H}) = \sum_{s=1}^N \sum_{t=1}^{T^s} \sum_{i=1}^{L_{s,t}} \biggl( | f^\text{shape}(\vect{x}^c_i, \vect{h}^s | \phi^\text{shape}) | + \\
    \lambda_3 |\nabla_\vect{x} f^\text{shape}(\vect{x}^c_i, \vect{h}^s | \phi^\text{shape}) - \vect{n}_i^c|_2
    \biggr),
\end{multline}
where $\vect{n}_i^c$ is the normal obtained by canonicalising the normal $\vect{n}_i$, along with the point $\vect{x}_i$ as described in Eq.~\ref{eq:canonicalization}, and $\nabla_\vect{x}$ denotes the spatial derivative. We compute the normal $\vect{n}_i$ on the point cloud using~\cite{newcombe2011kinectfusion}.
The term $E_\text{eik}(\cdot)$~\cite{gropp2020igr} enforces that the SDF prediction on the canonical surface should be zero and its derivative, i.e. normal direction, should match the canonicalised normal:
\begin{equation}
    E_\text{eik}(\phi^\text{shape}, \set{H}) = \sum_{s=1}^N \sum_{t=1}^{T^s} \sum_{i=1}^{L_{s,t}} \biggl(
    |\nabla_\vect{x} f^\text{shape}(\vect{x}^c_i, \vect{h}^s | \phi^\text{shape})|_2 - 1
    \biggr)^2.
\end{equation}

\noindent\textbf{Insights.}
Our objective $E^{shape}(\phi^{shape}, \set{H})$ allows us to fuse all partial canonical frames into a single continuous shape for each subject, averaging out the pose-dependent artefacts. 
The subject-specific geometry of the canonical shape can be encoded in their respective latent codes $\vect{h}^s$, whereas the decoder can freely learn common information across subjects. 

\subsection{NSF for Pose-Dependent Deformation}
\label{sec:manifold_implicit_function}
\noindent\textbf{Neural Surface Deformation Field.}
In the previous Section we described how to learn a pose-independent fusion shape by fusing input observations. 
But to faithfully reproduce the detailed 3D shape of a person we need to model fine-grained pose-dependent deformations.
Leveraging the NSF introduced in Sec.~\ref{sec:nsf}, we define a deformation field on the top of the fusion shape surface $\set{B}^s$:
\begin{equation}
f_{\phi}: \mathcal{S}^{2} \subset \mathbb{R}^3 \rightarrow \mathbb{R}^3,
\end{equation}
where points on the surface $\mathcal{S}^{2}$ are mapped to their corresponding pose-dependent displacements $\mathbb{R}^{3}$ in the canonical space.
Similar to our fusion shape, our deformation fields are also parameterized by a combination of subject-specific latent codes $\set{F}=\{\mat{F}^s\}_{s=1}^N$, and a pose conditioned decoder network $f^\text{pose}(\cdot|\phi^\text{pose})$. More specifically, the deformed points for the subject $s$ is computed as:
\begin{equation}
    \mat{X}^p = \mat{X}^c + f^\text{pose}(\mat{F}^s(\mat{X}^c), \theta | \phi^\text{pose}),
    \label{eq: nsf_equation_application}
\end{equation}
where $\mat{F}^s(\vect{x}^c)$ denotes the latent feature queried at point $\vect{x}^c$ for subject $s$ and $\theta$ denotes the pose feature encoded by a MLP.
Our key idea is to learn a NSF for deformation directly and solely on the surface of the implicit fusion shape $\set{B}^s \subset \mathbb{R}^3$ for each subject. This requires addressing two key challenges: \emph{how to learn features $\mat{F}^s(\cdot)$ on the surface$?$} and \emph{how to handle off-surface query points for prediction$?$}. 

\noindent\textbf{Feature Learning On Surface.}
Volumetric and pixel-aligned implicit feature learning methods~\cite{bhatnagar2020ipnet, chibane20ifnet, saito2019pifu,saito2020pifuhd} learn features at regular grid locations and use bi-/tri-linear interpolation to compute features at intermediate points. We devise a similar strategy to learn features on a surface.
We first discretize the implicit fusion shape $\set{B}^s$ by Marching Cubes~\cite{LorensenC87mcubes} to extract an explicit surface.
Moreover, if the garments can be represented by SMPL~\cite{loper2015smpl} topology, we fit the SMPL+D model by minimizing the SDF value of SMPL vertices. 
The same explicit mesh topology allows us to quickly initialize feature space across different subjects. 
We use the vertices ($5,000 \sim 7,000$) on this surface to form the feature basis location of our surface. 
The features are learnt via an auto-decoder during training. 
The feature $\mat{F}^{s}(\vect{x}^c)$ at arbitrary surface point $\vect{x}^c \in \set{B}^s$ is obtained using barycentric interpolation between three nearest neighbours among the sampled basis points.
Our feature learning on surface is compact and unlike the 1D vectors retains 3D spatial arrangement. 
In addition, it is memory-efficient, whereas volumetric latent features~\cite{chibane20ifnet,bhatnagar2020ipnet,chibane2020ndf} at 128 resolution require learning $128^3 \sim2$mil. features, while we only need to learn about $7$k. features using a neural surface space. 
Our experiments demonstrate that learning a deformation field on a surface produces better results than volumetric and other competing feature learning approaches with significantly lower number of features.

\noindent\textbf{Projecting Off-surface Points Onto Surface.}
Feature learning on surface is quite straightforward and intuitive as described above.
But it requires the query point $\vect{x}^c$ to lie on the surface $\set{B}^s$ as the NSF is not even defined outside in $\mathbb{R}^3$. 
This is challenging because the canonical point $\vect{x}^c$ obtained by canonicalising the input observation $\vect{x}$ (Eq.~\ref{eq:canonicalization}) is pose-dependent and does not lie on the surface.
To this end we use a simple method to project off-surface canonical point to $\set{B}^s$~\cite{chibane2020ndf,tiwari2022posendf}. 
We use our pre-trained auto-decoder in Sec.~\ref{sec:fusion_shape} to obtain the SDF corresponding to the canonical point, and the gradient of this SDF gives us the normal direction perpendicular to the surface. 
We can use this to find the canonical surface point $\vect{x}^{cc}$ corresponding to $\vect{x}^{c}$.
\begin{equation}
    \vect{x}^{cc} = \vect{x}^{c} + f^\text{shape}(\vect{x}^{c}, \vect{h}^s | \phi^\text{shape}) \nabla_{\vect{x}_c} f^\text{shape}(\vect{x}^{c}, \vect{h}^s | \phi^\text{shape}).
\end{equation}
With this surface projection we can obtain the correspondence $\vect{x}^{cc}$ on the fusion shape of each pose-dependent canonical point $\vect{x}^{c}$.
Afterwards, we can lift the neural surface feature from $\vect{x}^{cc}$, $\mat{F}^s(\vect{x}^{c}) \gets \mat{F}^s(\vect{x}^{cc})$.

\subsection{Self-supervised Cycle Consistency}
\label{sec:cycle_consistency}
\noindent\textbf{Reposing via Skinnning.} 
Once we obtain the pose-dependent deformation on top of the fusion shape in~\ref{eq: nsf_equation_application}, we use standard linear blend skinning~\cite{lewis2000pose}, to repose points:
\begin{equation}
  \mat{X}^{pp} = \left(\sum_{i=1}^{K}w_i(\mat{X}^p)\mat{T}_i(\theta)\right)\mat{X}^p,   
\end{equation}
where $\mat{X}^p = \{\vect{x}^p_i\}_{i=1}^{L_t}$ is the NSF predicted pose-dependent canonical points and $\mat{X}^{pp} = \{\vect{x}^{pp}_i\}_{i=1}^{L_t}$ is the reposed pose-dependent points. 
Note that $\mat{X}^{pp}$ can be considered as the reconstruction of input observation $\mat{X}_t$.

\noindent\textbf{Self-supervised Learning.} 
The NSF, namely subject-specific surface features $\set{F}=\{\mat{F}^s\}_{s=1}^N$ together with the pose-conditioned decoder network $f^\text{pose}(\cdot|\phi^\text{pose})$ can be trained end-to-end by ensuring that our posed reconstruction $\mat{X}^{pp}$ matches the input point cloud $\mat{X}_t$.
This can be formulated as the following self-supervised objective:
\begin{multline}
    \label{eq:loop_closure}
    E^\text{pose}(\phi^\text{pose}, \set{F}) = 
    \sum_{s=1}^N \sum_{t=1}^{T^s} \sum_{i=1}^{L_{s,t}}  
    \bigg( |\vect{x}_i - \vect{x}^{pp}_i|_2 + |\vect{n}_i - \vect{n}^{pp}_i|_2 \\
    + d^\text{CD}(\vect{x}_i, \vect{x}^{pp}_i) +
    E^\text{pose}_\text{reg}
    \bigg),
\end{multline}
\begin{equation}
    E^\text{pose}_\text{reg} = |\vect{x}^p_i - \vect{x}^c_i|_2 + |\mat{F}^s(\vect{x}^c_i)|_2 + EDR(\vect{x}_i^c),
\end{equation}
where $EDR(\vect{x}^c_i) = |\mat{F}^s(\vect{x}^c_i) - \mat{F}^s(\vect{x}^c_i + \omega))|_2$ and $\omega$ is random small scalar. 
$d^\text{CD}(\cdot, \cdot)$ denotes uni-directional Chamfer distance.
Eq.\ref{eq:loop_closure} forces that the predicted skinned points ($\vect{x}^{pp}_i$) and corresponding normals ($\vect{n}^{pp}_i$) match the input posed points ($\vect{x}_i$) and their normals ($\vect{n}_i$). The regularisation term $E_\text{reg}^\text{pose}$ contains an L2 regulariser on the deformation field and neural surface feature as well as EDR term~\cite{shue2022triplanediff} which enforces spatial smoothness on the feature space.

\subsection{Inference and Surface Extraction.}
\label{sec:surface_extraction}
At the inference time, we predict the pose-dependent deformation for vertices $\mat{V}^c$ of our base fusion shape $\set{B}^s$, and apply LBS~\cite{lewis2000pose} with given desired pose to obtain its location $\mat{V}^{pp}$ in the pose space. Because of the continuity of NSF, the fusion shape $\set{B}^s$ here can be discretized with arbitrary resolution and topology. We use the original edge connectivity on fusion shape $\set{B}^s$ and posed vertices $\mat{V}^{pp}$ to obtain the posed mesh, which ensures the coherency over different poses. Specifically for reconstruction task, where the partial point cloud is available, we freeze the deformation function $f^\text{pose}(\cdot)$ and fine-tune the neural surface feature via minimizing the single-directional Chamfer distance between the input partial shape and our reconstructed mesh together with the Laplacian smoothness loss~\cite{laplaciansmoothness2} of the reconstructed mesh. Our NSF guarantees the coherent direct mesh output at arbitrary resolution without performing expensive marching cubes as in~\cite{chen2022gdna,chen2021snarf, deng2019nasa, dong2022pina, Mihajlovic2021leap, tiwari21neuralgif, wang2021metaavatar} or Poisson reconstruction~\cite{lin2022fite, ma2022skirt, ma2021pop, zhang2023closet}

\begin{table*}[t!]
  \centering
\caption{\label{tab:recon}
We evaluate our method on the task of reconstructing 3D shape from monocular depth point clouds on
% Quantitative results of reconstruction from monocular depth, evaluated on rendered depth from the captured 
BuFF~\cite{zhang2017buff}, CAPE~\cite{ma2020cape}, and synthesized ReSynth~\cite{ma2021pop} data. Our method performs better than existing methods both quantitatively and qualitatively.
% Best results are in boldface.
}
\begin{tabular}{ l c c c c c c c c c  }

 \hline
\multirow{2}{*} {Method} & \multicolumn{3}{c}{BuFF Data~\cite{zhang2017buff}} & \multicolumn{3}{c}{CAPE Data~\cite{ma2020cape}} & \multicolumn{3}{c}{Resynth Data~\cite{ma2021pop}}\\
 \cline{2-10}
{} & {CD (cm) $\downarrow$} & {NC $\uparrow$} & IoU $\uparrow$ & {CD (cm) $\downarrow$}  &{NC $\uparrow$} & IoU $\uparrow$ & {CD (cm) $\downarrow$} & {NC $\uparrow$} & IoU $\uparrow$ \\
     
 \hline
 
% SMPL &     & &  & $1.49$ & $0.909$ & $0.762$ \\

 DSFN~\cite{burov2021dsfn}   & $1.56$    & $0.916$ & $0.832$ & - & - & - & - & - & -\\
 PINA~\cite{dong2022pina}   &  $1.10$   & $0.927$ & $0.879$ & $\textbf{0.62}$ & $0.906$ & $\textbf{0.941}$ & - & - & - \\

 \emph{Ours, w/o deformation}  &  $0.97$   & $0.922$ & $0.851$  & $0.86$ & $0.929$ & $0.869$ & $1.14$ & $0.915$ & $0.846$ \\
 \emph{Ours, complete} &  $\textbf{0.69}$   & $\textbf{0.930}$ & $\textbf{0.895}$  & $0.65$ & $\textbf{0.940}$ & $0.911$  & $\textbf{0.92}$ & $\textbf{0.917}$ & $\textbf{0.887}$ \\
 \hline
\end{tabular}
\end{table*}

\section{Experiments}\label{sec:experiments}

\paragraph{Datasets.}
We evaluate the results of our method qualitatively and quantitatively on single-view point cloud obtained from monocular depth sequences.
We rendered the depth sequences from the BuFF~\cite{zhang2017buff, ponsmoll2017clothcap} dataset and the CAPE~\cite{ma2020cape, ponsmoll2017clothcap} dataset using Kinect camera parameters, same as our baselines~\cite{burov2021dsfn, dong2022pina} and unproject monocular depth to use as our input along with the SMPL poses.
For real data, we use Kinect depth sequences provided in DSFN~\cite{burov2021dsfn}. 
We experiment with loose garments like skirts from the Resynth~\cite{ma2021pop, ma2022skirt} dataset.

\paragraph{Metrics.} To evaluate the error of our method we will rely on Chamfer distance (in $cm$), the normal correctness, and the IoU between the ground-truth mesh and the reconstructions of our body model. The formulation of our metrics can be found in supp. material. These evaluation metrics are also applied to our baselines~\cite{burov2021dsfn, dong2022pina}.

\paragraph{Baselines.} The work closest to ours is PINA~\cite{dong2022pina} as they have the same problem setting. 
% Code for PINA is not available but authors provided us with pre-computed results and we trained our model using the same data and settings. 
DSFN~\cite{burov2021dsfn} is another baseline that uses neural network to learn SMPL-based 3D avatars from monocular RGB-D video. 
Since the code of PINA and DSFN is both not released, we train our model using the same data and compare with the pre-computed results provided by authors.
We also compare with POP~\cite{ma2021pop}, MetaAvatar~\cite{wang2021metaavatar}, and NPMs~\cite{palafox2021npms} on CAPE~\cite{ma2020cape, ponsmoll2017clothcap} dataset. Here, we modify the Chamfer distance in POP~\cite{ma2021pop} to unidirectional, allowing it accept single-view point cloud as input. 
Apart from these recent works, we also deploy a simple yet intuitive baseline: posing the naked SMPL shape and our learned fusion shape (w/o NSF). These baselines highlight the importance of learning pose-dependent deformations in NSF.
% SMPL baselines require careful initialisation~\cite{openpose, guler2018densepose, ianina2022bodymap, xiu2022econ} without which they get stuck in local minima.

\begin{figure*}[t]
\begin{center}
    \centering
    \captionsetup{type=figure}
\includegraphics[width=\textwidth]{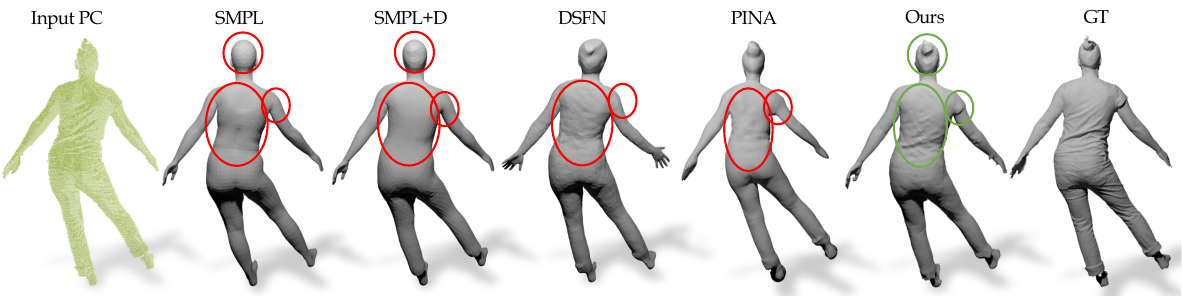}
\includegraphics[width=\textwidth]{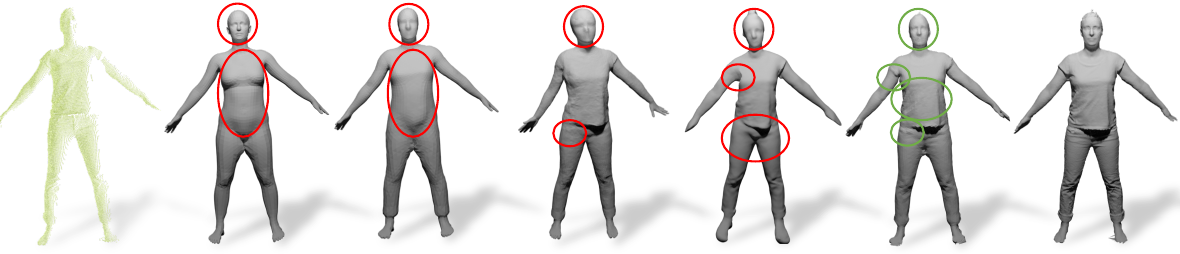}
\includegraphics[width=\textwidth]{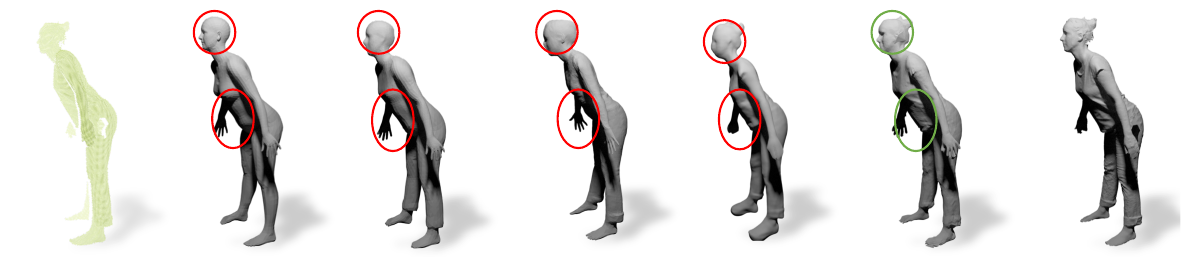}
\caption{\label{fig:recon}
Partial point cloud reconstruction on BuFF~\cite{zhang2017buff}: We first compare with fitting SMPL and SMPL+D models to our partial point clouds and then compare against more contemporary baselines DSFN~\cite{burov2021dsfn} and PINA~\cite{dong2022pina}. Our method reconstructs more detailed avatars.}
\end{center}%
\end{figure*}

\subsection{Reconstruction Comparison with Baselines.}

\begin{figure*}[h]
\begin{center}
    \centering
    \captionsetup{type=figure}
\includegraphics[width=\linewidth]{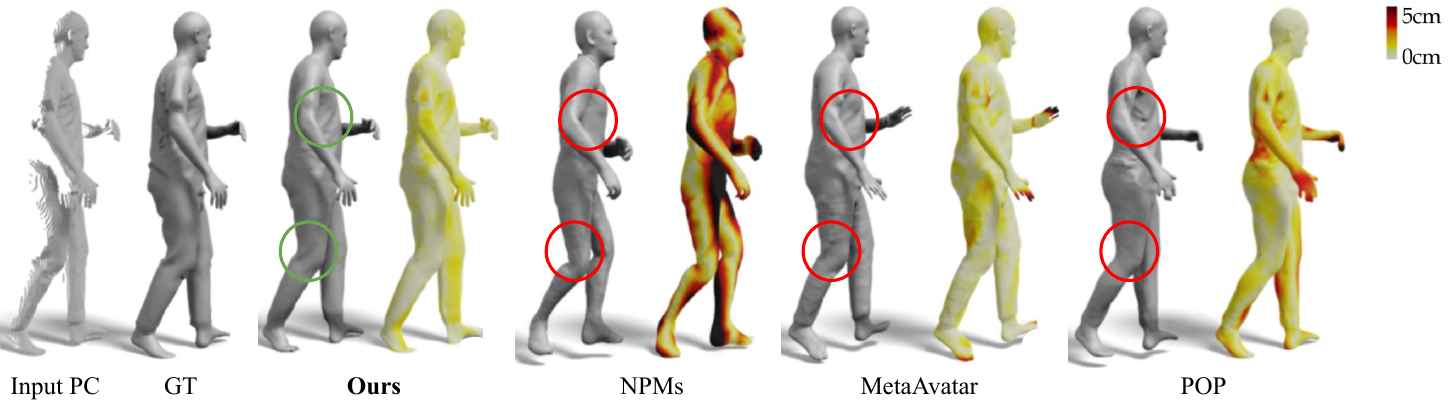}
\caption{\label{fig:cape_comparison}
Partial point cloud reconstruction on CAPE~\cite{ma2020cape, ponsmoll2017clothcap}: We compare with baselines NPMs~\cite{palafox2021npms}, MetaAvatar~\cite{wang2021metaavatar}, POP~\cite{ma2021pop} and visualize the reconstruction error on the surface. Our method achieves better reconstruction quality on this dataset.}
\vspace{-0.5cm}
\end{center}%
\end{figure*}

We test our method on the task of partial point cloud reconstruction. Given a sequence of a monocular point cloud, our goal is to recover a full clothed body model. Results are reported in Tab.~\ref{tab:recon} and Fig.~\ref{fig:recon},~\ref{fig:cape_comparison}. The results for each individual outfit of our method can be found in supp. material.
While the competing approaches~\cite{burov2021dsfn, dong2022pina} train a neural network per-subject, our method which is trained across multiple subjects, produces more reliable reconstructions with far less computational resources. Most essentially, our approach can reconstruct a sequence of coherent meshes at arbitrary resolution without retraining as in Fig.~\ref{fig:teaser}, which is not achievable by any of our baselines.

\subsection{Efficiency of Neural Surface Field.}
For this experiment we train 3 variants of our method with same neural networks and data but using three different feature representations, \ie volume~\cite{chibane20ifnet}, tri-plane~\cite{shue2022triplanediff} and neural surface features. We report our results in Tab.~\ref{tab:compare_feature}. 
% and Fig.~\ref{fig:compare_feature}. 
Our key idea to learn a deformation field on a neural surface is powerful and we can achieve better quality results with $10-100$x less learnable features compared to volumetric and tri-plane features. % Please see supp. mat. for qualitative examples.

Moreover, by avoiding per-frame surface extraction, NSF achieves from  $\sim40$x to $\sim 180$x faster compared to competitors at inference time. Please refer to supp. mat. for more detail.

\begin{table}[t]
  \centering
\caption{\label{tab:compare_feature}We compare our neural surface feature learning with existing volumetric~\cite{chibane20ifnet}, and tri-plane~\cite{shue2022triplanediff} feature representation. We show that we require significantly lower learnable parameters and produce better results.}
\resizebox{\columnwidth}{!}{
\begin{tabular}{ l r c c c }
 \hline
\multirow{2}{*} {Method}   & \multicolumn{4}{c}{BuFF Data~\cite{zhang2017buff} - Subject 00032} \\
 \cline{2-5}
{} & $\#$ Features &{CD $\downarrow$} & {NC $\uparrow$} & IoU $\uparrow$ \\
 \hline
Volume   & $262,144$   & $0.77$ & $0.925$ &  $0.884$ \\
Triplane    & $49,152$ & $0.74$ & $0.924$  & $0.885$ \\
 \emph{Ours, NSF}  & $\mathbf{6,890}$  & $\mathbf{0.66}$ & $\mathbf{ 0.928}$  & $\mathbf{0.899}$\\
 \hline
\end{tabular}
}
\end{table}

\begin{table}[t]
  \centering
    \caption{\label{tab:fine_tune}Our feature decoupling allows us to use our pre-trained network and quickly learn new subject specific features with little data and time. We show that in 10 mins, and by just using 10 frames (A) from a sequence, our model achieves similar performance as training on all the frames in 10 hrs (B).
    % Train on three subjects, ablate results for train and fine tune features for unseen subjects.
    }
    \resizebox{\columnwidth}{!}{
    \begin{tabular}{ l r r c c c }
         \hline
        \multirow{2}{*} {Operation}   & \multicolumn{5}{c}{BuFF Data~\cite{zhang2017buff} - Subject 00114} \\
         \cline{2-6}
        {} & $\#$ Frames & Time &{CD $\downarrow$} & {NC $\uparrow$} & IoU $\uparrow$ \\
         \hline
        (A) Train   & 126  & $\sim600'$ & $0.80$  & $0.929$  &  $0.881$  \\
        % Our, fine tune    &  &  &   & \\
        (B) Fine tune  & $10$  & $\sim 10'$ & $0.87$  & $0.907$  & $ 0.870$\\
         \hline
    \end{tabular}
    }
\end{table}

\subsection{Importance of Feature Fecoupling: Learning a New Avatar with 10 images in under 10 mins.}
\begin{figure}[h]
\begin{center}
    \centering
    \captionsetup{type=figure}
\includegraphics[width=\linewidth]{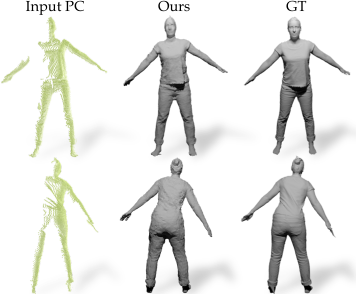}
\caption{\label{fig:new_avatar}
Point cloud reconstruction results: We learn the body model of a new subject given 10 frames in under 10 mins.}
\vspace{-0.5cm}
\end{center}%
\end{figure}

Our baselines~\cite{burov2021dsfn, dong2022pina} require training a new neural network for each subject. This is both computationally and data expensive.
Our method decouples generalizable neural networks and subject-specific features, and hence we can quickly learn new subject-specific features with small amounts of data, (\ie 10 depth images) in a short time (\(<10\)mins). Training a full neural network on the other hand requires several hours (see Tab.~\ref{tab:fine_tune}). We use 3 subjects from BUFF dataset for training and use 10 random frames from the \(4^{th}\) unseen subject for learning the body model. 
Our qualitative results in Fig.~\ref{fig:new_avatar} show that our decoupling allows us to learn models of new subjects easily with small amounts of data. 
Competing baselines~\cite{burov2021dsfn, dong2022pina} lack such capabilities, although their code is not available for fair comparison.
In our supplementary material, we also show that the generalizable decoder achieves superior performance compared to subject-specific decoder training.

\subsection{Animating Learnt Avatars.} 
\begin{figure*}[t]
\begin{center}
    \centering
    \captionsetup{type=figure}
\includegraphics[width=0.8\linewidth]{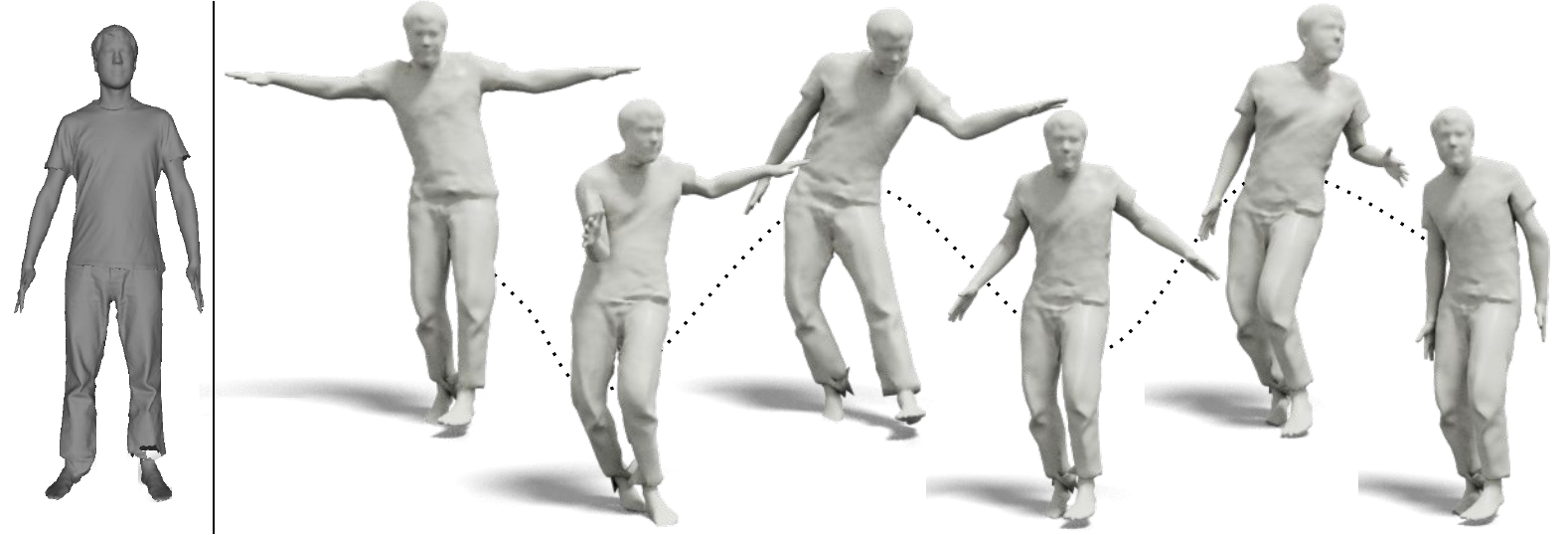}
\caption{\label{fig:animation}
Since our method learns a body model of the subject, we can use this model for re-animation. We show a reference scan of a person (left) and re-posed avatars of the subject (right). Note that NSF can directly output coherent animated meshes at arbitrary desired resolution (as in Fig.~\ref{fig:teaser}) without retraining, which is more flexible compared to state-of-the-art works. }
\vspace{-0.5cm}
\end{center}%
\end{figure*}
Our method can be efficiently used to manipulate the learnt model to unseen poses. This can be done by providing the desired input pose parameters to our method. We use our model trained on BUFF~\cite{zhang2017buff} and animate it with poses from AIST dataset~\cite{aist-dance-db}. 
Fig.~\ref{fig:animation} shows our learnt avatars in different poses. See supp. video and pdf for more examples.

\subsection{Results on Real Data.} 
\begin{figure}[h]
\begin{center}
    \centering
    \captionsetup{type=figure}
\includegraphics[width=0.9\linewidth]{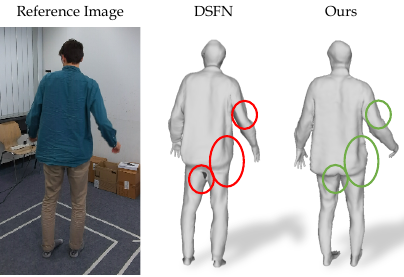}
\caption{\label{fig:real}
Generalization to real data: We show qualitative comparison with DSFN~\cite{burov2021dsfn} on their dataset captured using a Kinect. Our model generates more details and less artefacts. Note that the reference RGB image is not used in training.}
\vspace{-0.5cm}
\end{center}%
\end{figure}

In this experiment we test the generalization capability of our method on real data~\cite{burov2021dsfn}. Fig.~\ref{fig:real} demonstrates one example on real dataset from DSFN~\cite{burov2021dsfn}. Both the methods are trained using same data and we our method clearly outperforms the baseline. Please supp. mat. for more examples.

\subsection{Learning Textured Avatars.} 
\begin{figure}[h]
\begin{center}
    \centering
    \captionsetup{type=figure}
\includegraphics[width=\linewidth]{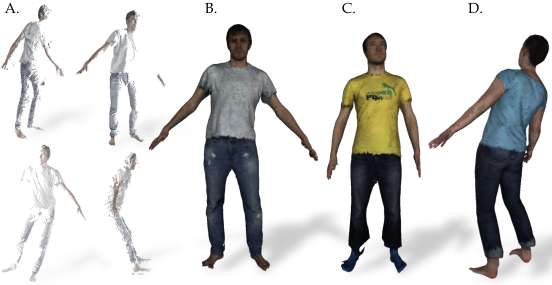}
\caption{\label{fig:texture}
We can learn textured 3D avatars of people from input partial point clouds. We show sample partial inputs (A) and corresponding learnt model (B) We show more avatars in C,D.}
\end{center}%
\end{figure}

We build our fusion shape by fusing multiple monocular point clouds and our canonicalization procedure ensures that we have explicit correspondence between the input posed space and the fusion shape. This allow us to directly lift the texture from the input point cloud onto the canonical shape and we obtain a textured body model of a person. Our baselines~\cite{burov2021dsfn, dong2022pina} have not shown such capabilities.
Fig.~\ref{fig:texture} shows examples of our learnt textured avatars.

\section{Conclusion}\label{sec:conclusions}

We introduced \emph{Neural Surface Fields} (\keyfeature): efficient, fine-grained manifold-based continuous fields for modeling articulated clothed humans. \keyfeature is capable of reconstructing meshes with arbitrary resolution without retraining while maintaining mesh coherency. 
NSF eliminates the expensive per-frame surface extraction, is about $40$ to $180$ times faster at inference time compared to baselines.  
\keyfeature is compact and preserve the 3D structure of the underlying manifold. 
NSF also enables applications like texture transfer and fine-tuning to adapt to a new subject. 
Our evaluation on rendered and captured data demonstrate the efficiency and the power of our proposed \keyfeature. 
We believe \keyfeature can lead to both real-world applications and useful tools for the 3D vision community. The code as well as models are available at \href{https://yuxuan-xue.com/nsf}{https://yuxuan-xue.com/nsf} for research purposes.

{\footnotesize
\paragraph{Acknowledgements}
We appreciate Y. Xiu, G. Tiwari, H. Feng, Y. Feng for their feedbacks to improve the work. 
This work is made possible by funding from the Carl Zeiss Foundation. 
This work is also funded by the Deutsche Forschungsgemeinschaft (DFG, German Research Foundation) - 409792180 (EmmyNoether Programme, project: Real Virtual Humans) and the German Federal Ministry of Education and Research (BMBF): Tübingen AI Center, FKZ: 01IS18039A. 
The authors thank the International Max Planck Research School for Intelligent Systems (IMPRS-IS) for supporting Y.Xue.
G. Pons-Moll is a member of the Machine Learning Cluster of Excellence, EXC number 2064/1 – Project number 390727645. 
R. Marin has been supported by Alexander von Humboldt Foundation Research Fellowship and partially from the European Union’s Horizon 2020 research and innovation program under the Marie Skłodowska-Curie grant agreement No 101109330.
}
{\small
\bibliographystyle{ieee_fullname}
\bibliography{References}
}

\end{document}